# Leveraging CNN and IoT for Effective E-Waste Management

Ajesh Thangaraj Nadar
*Department of Computer Engineering*
*SIES GST*
Nerul, India
ajeshraj402@gmail.com

Gabriel Nixon Raj
*Department of Computer Engineering*
*SIES GST*
Nerul, India
gabriel.nixonraj@gmail.com

Soham Chandane
*Department of Computer Engineering*
*SIES GST*
Nerul, India
szsohamchandane@gmail.com

Sushant Bhat
*Department of Computer Engineering*
*SIES GST*
Nerul, India
sushantbhatt1332@gmail.com

*Abstract*—In this day and age the number of electronic devices are growing exponentially due to which there is a problem of e-waste management. The tendency of just throwing out garbage especially electronic circuits and devices has made recycling more complex. By using IoT we will segregate these e-waste into several items which can then be sent to e-waste recyclers so they can recycle these items effectively. The proposed system will contain a camera and a weighing scale. by using object detection the e-waste will be segregated by their basis of components such as circuit boards ,sensors, wired etc. The weighing scale will determine the average price of the component with respect to its weight.

## I. Introduction

The increasing proliferation of electronic devices in the modern era has led to a significant surge in electronic waste, commonly known as e-waste. There are serious environmental and health risks associated with the incorrect dumping and insufficient recycling of e-waste. Therefore, the need for creative ways to efficiently handle and recycle e-waste is crucial. In order to solve the difficulties in managing e-waste, this article suggests an IoT-enabled system that streamlines the segregation and recycling procedures.

### A. E-waste management challenges

E-waste management is faced with a number of difficulties, such as the growing amount of e-waste generated by the increased use of electronic devices, the difficulty of recycling different components, insufficient infrastructure and resources, international trade and illegal dumping, concerns about data security and privacy, a lack of awareness and education, and the shift to a circular economy. To ensure correct disposal, recycling, and sustainable waste management practices, overcoming these issues calls for coordinated efforts and creative solutions.

### B. Growing E-waste accumulation

The growing amount of e-waste in Mumbai is currently causing the city to struggle. Urban centres that embrace technological breakthroughs have seen a considerable increase in the use of electronic gadgets, which has in turn caused a corresponding rise in the disposal of old or damaged electronic equipment. E-waste, which includes old cell phones, laptops, computers, and other electronic gadgets, contains dangerous materials that, if not handled carefully, pose serious threats to the environment and human health. Unfortunately, the incorrect disposal of electronic devices has led to their accumulation in landfills, creating environmental degradation and potential harm to the ecosystem. This has been made possible by the lack of efficient e-waste management methods. Local government, industry, and the community must give ethical and sustainable e-waste recycling practises top priority in order to combat this growing problem. In the face of this difficult situation, Navi Mumbai may do this to protect its environment and assure the welfare of its citizens.

### C. Importance of E-waste management

In today's digital world, the significance of e-waste management cannot be emphasised. The production of electronic garbage, or e-waste, has become an urgent worldwide problem due to the quick development of technology and the rising use of electronic gadgets. E-waste is made up of outdated electronic products including televisions, laptops, cell phones, and other electronic devices that frequently include dangerous metals like lead, mercury, and cadmium. E-waste improperly disposed of poses serious threats to the environment and human health since these harmful components can seep into the ground and water, destroying ecosystems and risking human health.

For these negative consequences to be reduced, effective e-waste management is essential. In addition to preventing environmental damage, proper e-waste recycling and disposal may help electronic gadgets retain important resources like rare earth elements and precious metals, which will lessen the need for new resource extraction. Additionally, a well run e-waste recycling system may foster employment growth and boost the economy of the recycling sector.

TABLE 1    E-WASTE GENERATION

| Year | E-waste generation (million metric tonnes) |
|---|---|
| 2015 | 1.97 |
| 2016 | 2.22 |
| 2017 | 2.53 |
| 2018 | 2.86 |
| 2019 | 3.23 |

## II. Statistics

E-waste production was projected by Balde et al. (2015) to have reached 41.8 million tonnes globally in 2014, and by 2018 it was predicted to reach 50 million tonnes. The expected annual increase rate for the e-waste stream is extremely large, ranging from 3% to 5%, claim Cucchiella et al. (2015). This rate is significantly greater than that of other waste streams, claim Singh et al. (2016).Abbreviations and Acronyms

TABLE 2  E-WASTE IN DIFFERENT CATEGORIES

| Categories | Amount (in million tonnes) |
|---|---|
| Temperature exchange equipment | 7.0 |
| Screens & monitors | 6.3 |
| Lamps | 1.0 |
| Large Equipment | 11.8 |
| Small equipment | 12.8 |
| Small IT & telecommunication equipment | 3.0 |

GRAPH 1  E-WASTE IN YEARS
SOURCE  ANNUAL REPORT MEITY

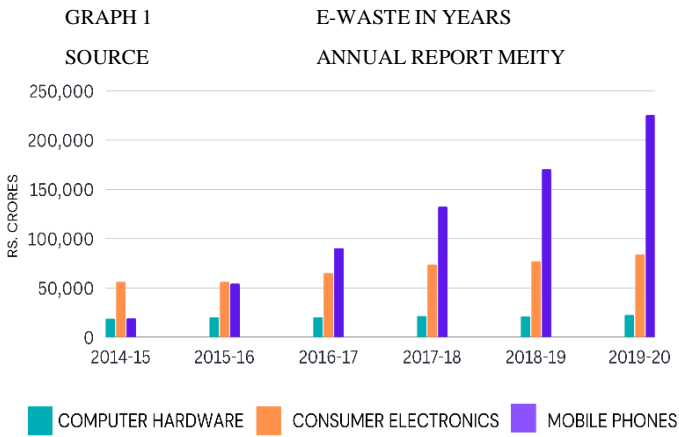

The primary sources of electronic waste include temperature exchange equipment, screens/monitors, small and big equipment, and other objects, according to which classifies electronic waste into a variety of categories. The fact that solar panels are evolving into a new kind of e-waste is also noted. The global PV waste stream is anticipated to be between 43,500 and 250,000 metric tonnes by the end of 2016, and this quantity is predicted to increase to 5.5 to 6 million tonnes by the year 2050, according to Weckend et al. (2016).

GRAPH 2  E-WASTE GENERATION IN INDIA
SOURCE  CSE, 2020

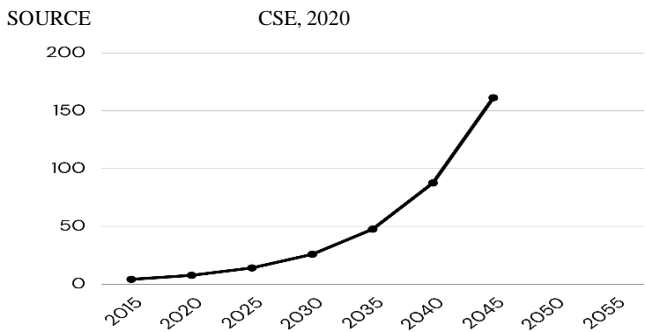

## III. Methodology & Algorithm

FLOWCHART 1  FLOWCHART OF WORKING MODEL

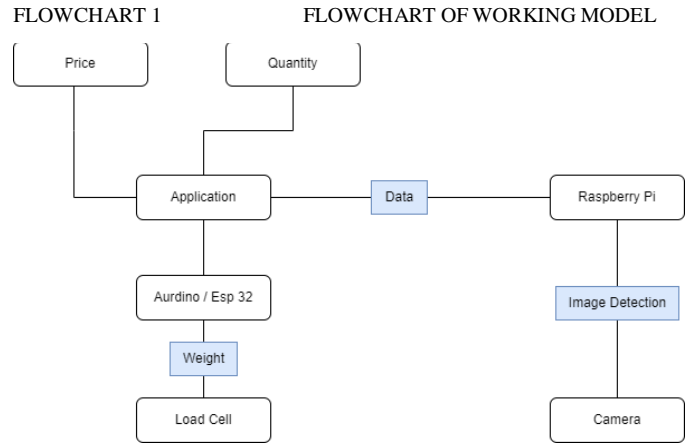

### A. Data preprocessing

In this study, we provide a detailed description of the data collection, annotation, and preprocessing steps required to produce an effective e-waste identification system. Utilising the coco pretrained model, Nonso, Barrowclough, Joseph, and Procter, Jack (2022) "Waste Classification Dataset" available on Mendeley Data, our objective is to produce a diverse and well-structured dataset to train a deep learning model for identifying and classifying various e-waste objects in real-time. A range of images from various websites that featured e-waste objects were collected throughout the data collection stage. The coco pretrained model. Waste Classification Dataset was produced by combining photographs taken with digital cameras of various electronic devices. The inclusion of a variety of electrical devices ensures that the model is trained on a representative collection of e-waste photographs. To allow accurate model training, the dataset was annotated using the VGG Image Annotator (VIA). Each image was meticulously annotated with bounding boxes that included certain e-waste items. The deep learning model uses these annotations as ground truth labels to precisely detect and classify various forms of e-waste. The dataset was then preprocessed to enhance model performance. In order to ensure uniform input ranges for the neural network during training, all photographs were resized to the same resolution and the pixel intensity was normalised. Using data augmentation methods like random flips and rotations, the dataset's variety was boosted, resulting in a robust and flexible training set.

In order to analyse the model effectively, the annotated dataset was split into training and testing sets using a stratified approach. About 80% of the photographs were part of the training set, whereas 20% were part of the testing set. By using this partitioning method, it is feasible to test the model's generalisation capabilities on hypothetical e-waste data. In conclusion, utilising the coco pretrained model. Waste Classification Dataset, we have effectively curated a comprehensive dataset for training a deep learning model capable of e-waste identification. The combined efforts of data collection, annotation, and preprocessing have resulted in a high-quality dataset that serves as the basis for the development of an accurate and effective e-waste identification system.

## B. Deep learning model

In order to train the deep learning model for e-waste identification, we utilised two effective tools: the COCO dataset and the YOLO (You Only Look Once) architecture. The COCO dataset, known for its enormous collection of unique pictures, was used to train the model. The dataset must contain complex situations and a wide range of objects in order to properly recognise diverse e-waste goods in real-world settings.

The key aspect was how well the YOLO architecture handled real-time object identification. YOLO analyses photographs in a single pass, enabling speedy and accurate inference, making it well suited for real-time applications like e-waste identification. The model's ability to analyse images properly is especially useful when employing e-waste detection systems on low resource devices or in circumstances where rapid reflexes are required for effective garbage management.

The coco pretrained model. Waste Classification Dataset was annotated with bounding boxes to indicate the locations of e-waste objects, and the COCO dataset was utilised to fine-tune the YOLO model during training. By combining these datasets, the model was able to learn a range of characteristic patterns of e-waste goods, leading to the development of a more precise and reliable identification approach. While optimising the model's parameters during training, stochastic gradient descent (SGD) was utilised to minimise the loss function, which combines the object detection and classification objectives. Thanks to its multi-task learning technique, the model can concurrently recognise and categorise e-waste objects effectively.

The COCO dataset and the YOLO architecture were used to effectively train a deep learning model for e-waste identification. The model's ability to handle a range of e-waste objects in real-time demonstrates its potential for effective and useful waste management solutions. By correctly identifying and categorising e-waste goods, this technique may contribute significantly to ecologically friendly waste management practises and environmental preservation

## C. Convolutional Neural Networks (CNNs) for E-Waste Detection

Convolutional neural networks (CNNs) play a significant role in our method for identifying e-waste via deep learning. CNNs were created expressly for tasks involving pictures, therefore they are good at automatically learning complicated characteristics from photos. They employ convolutions to recognise patterns like edges, textures, and shapes by gathering both low-level and high-level information in order to identify e-waste products. The ability of CNNs to generalise and extract useful features from sparse e-waste data is improved by pre-training them on large datasets like COCO. Transfer learning from COCO, which uses previously gained knowledge to e-waste identification, significantly enhances their performance. CNNs are effective at processing high picture volumes, making them a good choice for real-time e-waste identification applications. In general, CNNs are essential to ensuring precise and effective e-waste detection, supporting environmentally friendly waste management techniques.

## D. COCO

In our e-waste identification algorithm, we make use of the COCO dataset in a number of key ways. From the COCO dataset, we first extract the relevant images and annotations of e-waste. Then, using this subset of data to refine a pre-trained CNN, we apply transfer learning to adapt it to e-waste identification. To improve generalisation, techniques for data augmentation are employed. A tuned CNN that has been augmented with layers for object detection and classification serves as the model's basis. The accuracy is assessed using mean average precision (mAP) and intersection over union (IoU) on a segmented e-waste sub-dataset. Hyperparameter tweaking and fine-tuning help to increase model performance. For practical waste management applications, the model, which is applied in real-time, estimates the existence and locations of e-waste, ensuring environmental preservation and sustainability.

## E. YOLO

We use the effective YOLO (You Only Look Once) architecture, which is renowned for its real-time item identification capabilities, into our e-waste detection model. YOLO just requires one pass to process an image, making it perfect for devices with limited resources and time-sensitive applications like e-waste identification. A pre-trained CNN that has been optimised on the COCO dataset is paired with the YOLO architecture to provide a robust and flexible model. With the use of this transfer learning strategy, the model's capacity to recognise and categorise e-waste objects properly is improved. To ensure reliable and effective e-waste identification, YOLO's multi-task loss function integrates both object detection and categorization objectives. Once trained, the YOLO-based model is used in real-time applications to forecast the existence of e-waste and its locations for environmentally friendly waste management techniques. A more ecologically conscious future is made possible by the seamless integration of YOLO in our e-waste detection system, which makes accurate and real-time identification of electronic trash possible.

## IV. HARDWARE

To take pictures of e-waste objects, connect the camera to the Raspberry Pi. Securely fasten the camera module to the Raspberry Pi's CSI port. Connect the Raspberry Pi to a power source, display (if desired), keyboard, and mouse to complete setup. For data transfer and software upgrades, make sure the internet is connected. In order to measure the weight of e-waste components, connect the load cell to the Arduino UNO/ESP32. The load cell is coupled to the Arduino UNO/ESP32, which serves as the controller for weight measurement, to be used for managing e-waste. The necessary pins on the Arduino UNO/ESP32 are connected to the signal wires of the load cell, which are commonly color-coded. The ground (GND) wire is linked to the GND pin, the power (VCC) wire to the 5V or 3.3V pin, and the signal wires to the analogue input pins.

The load cell's analogue voltage output can be read by the Arduino UNO/ESP32 after it is attached. The voltage readings can be transformed into weight measurements by using the proper scaling and calibration factors. These weight measurements offer useful information for a variety of e-waste management tasks, including inventory control, evaluating recycling effectiveness, and calculating the total weight of e-waste components. Informed decision-making and efficient management techniques to lessen the environmental impact of e-waste are made possible by accurate weight assessment utilising the load cell.

### A. Object detection

To implement e-waste management, one has to install SSD MobileNet v3, a pre-trained item detection model, on the Raspberry Pi. This procedure calls for purchasing the pertinent model files and setting up the Raspberry Pi with the essential libraries. The model will allow you to recognise and identify e-waste objects, such as circuit boards, sensors, and cables, in real-time using the recorded video input from the camera connected to the Raspberry Pi.

Initialise the Raspberry Pi's MQTT client after the model has been loaded in order to connect to the MQTT broker. Configure the mobile app and Raspberry Pi to connect to a MQTT broker, such as Mosquitto, that has been set up on the Raspberry Pi or another server. This makes communication and data sharing simple. You can communicate the names of the recognised items with other devices or apps that are subscribed to that subject by publishing the detected e-waste component information to a particular MQTT topic, enabling efficient e-waste management and additional analysis.

## V. SOFTWARE

### A. MQTT

To allow MQTT communication, install a MQTT broker, such as Mosquitto, on either the Raspberry Pi or another server with a recognised IP address. This broker will serve as the primary hub for message exchange. Configure the Raspberry Pi and the mobile app to connect to the MQTT broker, creating a communication channel between them. The Raspberry Pi may broadcast the recognised e-waste component name to a specific MQTT topic, which the mobile app can subscribe to, to ensure smooth data flow and enable real-time updates.

The Raspberry Pi can publish the names of the recognised e-waste component names to the chosen MQTT subject by setting up this MQTT communication configuration. The mobile app can then receive and process the published data since it is subscribed to the same subject. This enables effective management and decision-making in the field of e-waste management by enabling efficient tracking and monitoring of e-waste components.

### B. Blynk interface

Setting up a project in the Blynk app and acquiring the authentication token are the first steps towards utilising the Blynk platform for IoT. Install the Blynk library on the Raspberry Pi and use the authentication token for authentication to establish connectivity between the Raspberry Pi and Blynk. Create a user-friendly interface for the Blynk app that enables users to easily order recyclable e-waste components that are stacked on pallets with this foundation already in place. Real-time updates on the availability of e-waste components can be effortlessly transferred to the Blynk app by integrating MQTT communication between the Raspberry Pi and the app, ensuring users have the most recent information when placing their orders.

Each e-waste component should be fully described in the Blynk app's interface, including its weight, quantity, and anticipated cost, so that users may make informed judgements using the information supplied. The Blynk app simplifies the process of getting reusable e-waste components and delivers timely information, resulting in effective and sustainable e-waste management.

## VI. CONCLUSION

Electronic garbage (e-waste), which poses substantial environmental and health dangers if improperly managed, has significantly increased as a result of the fast expansion of electronic gadgets in the current digital age. In order to improve the segregation and recycling processes, an Internet of Things (IoT)-enabled system has been proposed as a creative solution to the problems connected with managing e-waste. E-waste management is faced with a number of obstacles, such as the increasing amount of e-waste produced, challenges with recycling different components, insufficient infrastructure and resources, illicit dumping, worries about data security, lack of knowledge, and the move towards a circular economy. To guarantee efficient disposal, recycling, and sustainable waste management practises, tackling these difficulties demands coordinated efforts and innovative solutions.

Finding practical solutions is urgently needed given the growth of e-waste in urban areas like Mumbai. Electronic waste has accumulated in landfills as a result of incorrect disposal, potentially harming ecosystems and deteriorating the environment. Local authorities, businesses, and communities must place a high priority on ethical and sustainable e-waste recycling practises in order to tackle this problem. In today's digital age, efficient e-waste management is crucial. E-waste poses serious risks to the environment and to people's health since it is made up of outmoded electronic items that contain dangerous elements. In addition to preventing environmental harm, proper e-waste recycling and disposal also helps protect priceless resources and promotes economic growth in the recycling industry.

According to statistics, there is an astounding quantity of e-waste produced worldwide, with an expected yearly growth rate of 3% to 5%. The various character of e-waste and the demand for effective management techniques are highlighted by classifying it into numerous categories, such as temperature exchange equipment, screens/monitors, lighting, big and small equipment, and tiny IT and communications equipment.

In conclusion, the suggested IoT-enabled system for managing e-waste provides a viable answer to the problems brought on by electronic trash. This approach improves sustainability, lowers environmental and health concerns, and aids in the shift to a circular economy by efficiently recognising, classifying, and recycling e-waste. To assure the adoption of moral and sustainable e-waste management practises on a worldwide scale, the implementation of this new method necessitates coordination among stakeholders, including governments, industry, and communities

Through these coordinated efforts, we can open the door for future generations to enjoy more prosperity and environmental awareness.